\documentclass[10pt,twocolumn,letterpaper]{article}

\usepackage{iccv}
\usepackage{times}
\usepackage{epsfig}
\usepackage{graphicx}
\usepackage{amsmath}
\usepackage{amssymb}

\usepackage{graphicx}
\usepackage{amsmath}
\usepackage{amssymb}
\usepackage{booktabs}
\usepackage{algorithm}
\usepackage{algorithmic}
\usepackage{multirow}
\usepackage{subfig}

\usepackage[pagebackref=true,breaklinks=true,letterpaper=true,colorlinks,bookmarks=false]{hyperref}

\usepackage[capitalize]{cleveref}
\crefname{section}{Sec.}{Secs.}
\Crefname{section}{Section}{Sections}
\Crefname{table}{Table}{Tables}
\crefname{table}{Tab.}{Tabs.}

\iccvfinalcopy 

\def\iccvPaperID{5449} 

\ificcvfinal\fi

\begin{document}

\title{Hard Adversarial Example Mining for Improving Robust Fairness}

\author{Chenhao Lin\\
Xi'an Jiaotong University\\
{\tt\small  linchenhao@xjtu.edu.cn}
\and
Xiang Ji\\
Xi'an Jiaotong University\\
{\tt\small xiangji@stu.xjtu.edu.cn}
\and
Yulong Yang\\
Xi'an Jiaotong University\\
{\tt\small xjtu2018yyl0808@stu.xjtu.edu.cn}
\and
Qian Li\\
Xi'an Jiaotong University\\
{\tt\small qianlix@xjtu.edu.cn}
\and
Chao Shen\\
Xi'an Jiaotong University\\
{\tt\small chaoshen@mail.xjtu.edu.cn}
\and
Run Wang\\
Wuhan University\\
{\tt\small wangrun@whu.edu.cn}
\and
Liming Fang\\
Nanjing University of Aeronautics and Astronautics\\
{\tt\small fangliming@nuaa.edu.cn}
}

\maketitle

\begin{abstract}
Adversarial training (AT) is widely considered the state-of-the-art technique for improving the robustness of deep neural networks (DNNs) against adversarial examples (AE). Nevertheless, recent studies have revealed that adversarially trained models are prone to unfairness problems, restricting their applicability.
In this paper, we empirically observe that this limitation may be attributed to serious \textit{adversarial confidence overfitting}, i.e., certain adversarial examples with overconfidence.
To alleviate this problem, we propose HAM, a straightforward yet effective framework via adaptive Hard Adversarial example Mining.
HAM concentrates on mining hard adversarial examples while discarding the easy ones in an adaptive fashion. 
Specifically, HAM identifies hard AEs in terms of their step sizes needed to cross the decision boundary when calculating loss value.
Besides, an early-dropping mechanism is incorporated to discard the easy examples at the initial stages of AE generation, resulting in efficient AT.
Extensive experimental results on \textcolor{black}{CIFAR-10, SVHN, and Imagenette} demonstrate that HAM achieves significant improvement in robust fairness while reducing computational cost compared to several state-of-the-art adversarial training methods. \textit{The code will be made publicly available.}
\end{abstract}

\section{Introduction}
\label{intro}

In the past decade, Deep Neural Networks (DNNs) have advanced rapidly, reaching a level beyond human intelligence in many areas.
However, several studies~\cite{Szegedy2014IntriguingPO, Goodfellow2015ExplainingAH} have discovered that when exposed to specifically designed imperceptible perturbations added to the original inputs, known as adversarial examples (AEs), the accuracy of DNNs can drop dramatically. 

Various approaches have been proposed to enhance the defense capabilities of DNNs against AEs.
Adversarial training (AT) has been demonstrated to be one of the most effective strategies~\cite{Madry2018TowardsDL}. Nevertheless, recent research \cite{Xu2021ToBR,Wang2021ImbalancedAT} have observed that the adversarially trained models usually suffer from a serious unfairness problem, i.e., there is a noticeable disparity in accuracy between different classes, seriously restricting their applicability in real-world scenarios. Although some solutions have been proposed, the average robustness fairness score is still low and needs to be urgently addressed.   
On the other hand, several recent studies \cite{Zhang2019YouOP, Shafahi2019AdversarialTF, Wong2020FastIB} have focused on achieving efficient adversarial training. They attempt to address the significantly high computational cost challenge by cutting down the iteration times of example generating or model training. Unfortunately, accelerating AT is often incompatible with the high fairness of the model and most existing fast adversarial training approaches \cite{Zhang2019YouOP, Shafahi2019AdversarialTF} reduce the computation time at the expense of sacrificing fairness or accuracy.

\begin{figure*}[!htb]
\centering
\includegraphics[scale=0.31]{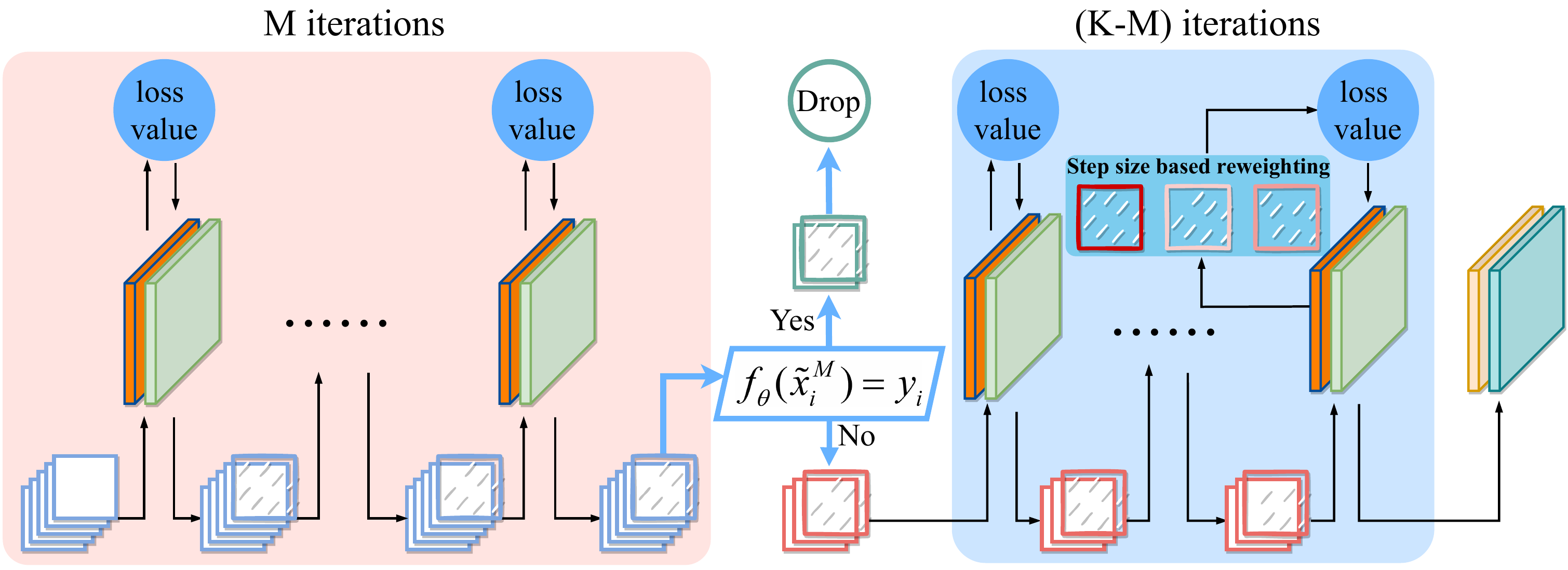} 
\caption{Pipeline of HAM. The orange and green blocks represent the original model while the straw yellow and mint green blocks represent the updated new model. The red and green squares represent hard and easy examples, respectively. The red square with darker colors represents an example with larger maximum step size and vice versa.}
\label{fig1}
\end{figure*}

In this paper, we aim to improve the fairness of adversarially trained models while reducing the high computational burden associated with AT. We propose to study AT from the perspective of sample mining, leveraging insights from the existing literature \cite{Shrivastava2016TrainingRO,Toneva2019AnES,Zhang2021GeometryawareIA} and our preliminary experimental analyses, in which the most important points are: (1) Compared to standard training, AT is a more challenging task than standard training, as it requires the model to prioritize minimizing the loss on robust examples rather than non-robust ones. This results in the overconfidence problem of AT models on certain AEs, which we refer to as \textit{adversarial confidence overfitting}. This phenomenon can lead to unsatisfying unfairness problems in AT. (2) The selection of appropriate adversarial examples for use in AT can be effective in mitigating the unfairness problem of standard AT. (See more details in Section \ref{Sec.3}).

To this end, this paper proposes an adaptive Hard Adversarial example Mining framework, dubbed HAM, to address the fairness and efficiency challenges in AT.
To make AT pay more attention to the hard examples rather than consistently minimizing the loss of correctly classified AEs, HAM adaptively mines hard AEs and discards the easy examples in terms of whether the AEs cross the decision boundary.
In addition, an early-dropping mechanism is proposed to drop the AEs that fail to cross the decision boundary during the first several stages of AEs generation, thus reducing the computational cost of AT.
Overall, our main contributions can be summarized as follows:

\begin{itemize}
\item We reveal the adversarial confidence overfitting problem in AT and propose HAM, an adaptive hard adversarial example mining framework to alleviate this issue, resulting in improved fairness of adversarially trained models.
\item We devise an early-dropping mechanism, which saves 46\% training time of traditional AT while preserving the robustness of the model.
\item The experimental results on three datasets demonstrate that our HAM significantly outperforms standard AT and state-of-the-art AT methods in terms of model robust fairness and computational cost with comparable adversarial accuracy.
\end{itemize}


\section{Related work}\label{Sec.2}
\subsection{Adversarial training}\label{Sec.2.1}

Among different types of existing defense strategies, such as AEs detection, model ensemble, and parameter randomization, adversarial training has proven to be the most successful defense against adversarial attacks \cite{Rice2020OverfittingIA}. The mainstream AT uses PGD to generate AEs, known as PGD-AT \cite{Madry2018TowardsDL}, and various methods have been proposed to improve its performance from different aspects. TRADES \cite{Zhang2019TheoreticallyPT} divides the original loss function into two parts, representing the accuracy for a clean sample and the robustness for malicious disturbance, by adjusting the hyperparameters. According to the entropy of its predicted distribution, Entropy Weighted AT scheme \cite{kim2021entropy} weighs the loss for each AE to increase the robustness accuracy.
\textcolor{black}{HAT \cite{Rade2022ReducingEM} proposes to generate an additional sample with a larger perturbation and give it an ``error" label as a helper for achieving better robustness.}
Despite their success, most existing approaches fail to address the issues of unfairness and high computational cost in AT. 

\subsection{Robust fairness} \label{Sec.2.2}
Recent studies have observed a serious fairness problem in AT, leading to a significant accuracy and robustness disparity between different classes of adversarially trained models. Sun et al. \cite{sun2022towards} show that the trade-off among fairness, robustness, and model accuracy can introduce a great challenge for robust deep learning. They propose a Fair and Robust Classification method by modifying the input data and models to address the fairness challenge. Xu et al. \cite{Xu2021ToBR} empirically find the serious deficiency of AT in fairness and attempt to mitigate this problem using the proposed Fair-Robust-Learning framework. 
More recently, Sun et al. \cite{sun2022improving} propose Balance Adversarial Training (BAT) to improve robust fairness by balancing the source-class and target-class fairness. Ma et al. \cite{ma2022tradeoff} theoretically study the trade-offs between adversarial robustness and class-wise fairness, and a fairly adversarial training method is proposed to mitigate the unfairness problem. Although several approaches have been proposed to study the robust unfairness problem in AT, the fairness of the adversarially trained models is still not satisfying and needs further improvement. 

\subsection{Efficient adversarial training} \label{Sec.2.3}
\textcolor{black}{To reduce the huge amount of computation consumed by multiple iterations of PGD during AT, several methods have been proposed \cite{Zhang2019YouOP, Zheng2020EfficientAT, ye2021amata}. Free AT \cite{Shafahi2019AdversarialTF}  proposes to recycle the gradient information when updating model parameters during AT to improve the AT efficiency. Fast AT \cite{Wong2020FastIB} finds that the specifically designed FGSM \cite{Goodfellow2015ExplainingAH} AT can achieve comparable performance with PGD AT and highly reduce the training time. They greatly shorten the adversarial training time, but at the expense of sacrificing part of the fairness and defensiveness of the models.}

\begin{figure}[!t]
\centering
\includegraphics[scale=0.55]{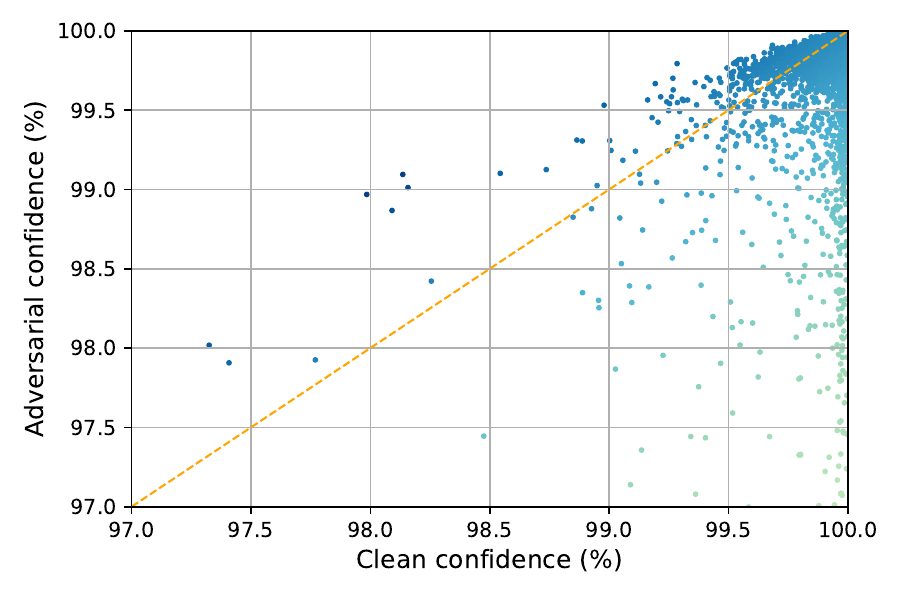} 
\caption{Confidence scores of adversarial examples and corresponding clean examples in adversarially trained WideResNet-34-10 \cite{Zagoruyko2016WideRN} on CIFAR-10 \cite{Krizhevsky2009LearningML} dataset. Darker points mean that it has much larger adversarial confidence than clean confidence. Brighter points represent examples with larger clean confidence than adversarial confidence.}
\label{fig2}
\end{figure}

\subsection{Hard negative mining}\label{Sec.2.4}
Hard negative mining technique is also related to our work. These methods \cite{girshick2014rich,he2015spatial} are widely applied in the field of object detection to solve the problem of imbalance in the proportion of positive and negative samples. It selects the RoIs that are difficult for model reasoning as representatives of all negative cases and thus resulting in a dramatic increase in test recalls. Online Hard Example Mining (OHEM) \cite{Shrivastava2016TrainingRO} method extends hard negative mining by using an online selection process with both hard negatives and hard positives, which reduces the training time considerably and improves the performance. The success of the hard example mining mechanism \cite{Lin2020FocalLF,Cui2019ClassBalancedLB} in various tasks motivates us to investigate adversarial training from the perspective of example mining.

\section{Methodology}\label{Sec.3}

\subsection{HAM overview}\label{Sec.3.1}
We consider a robust classification task with a given dataset $X = \{ {x_i},i = 1,2, \cdots ,N\} $ and a perturbation budget $\epsilon $. AT is expected to solve the following min-max optimization problem:
\begin{equation}\label{Eq.1}
\mathop {\min }\limits_{{f_\theta }} \sum\limits_{i = 1}^N {\mathop {\max }\limits_{{{\left\| {\tilde x_i^K - {x_i}} \right\|}_p} \le \epsilon} {\rm{ }}\mathcal{L}({f_\theta }(\tilde x_i^K),{y_i})} ,
\end{equation}
where ${f_\theta }( \cdot )$ is the DNN parameterized with $\theta $, ${y_i}$ is the true label corresponding to input ${x_i}$, $\mathcal{L}( \cdot )$ is the loss function. ${\left\| {\tilde x_i^K - {x_i}} \right\|_p}$ is the $l_p$-norm used to bound the adversarial perturbation ($p=2$ or $\infty$). $\tilde x_i$ is the adversarial example corresponding to clean example $x_i$. PGD-AT adopts a $K$-step PGD attack iterations process to obtain the $\tilde x_i^K$, which can be formalized as,
\begin{equation}\label{Eq.2}
\tilde x_i^{j + 1} = Cli{p_\epsilon}(\tilde x_i^j + \alpha  \cdot sign({\nabla _{\tilde x_i^j}}\mathcal{L}({f_\theta }(\tilde x_i^j),{y_i}))),
\end{equation}
where $\alpha $ is the step size of each iteration, and $Clip( \cdot )$ is the projection operation that guarantees $\tilde x_i$ is within the $l_p$-ball.

The intuition of the HAM is to pay more attention to the hard AEs in the adversarial training procedure, whose training objective can be formalized as:
\begin{equation}\label{Eq.3}
\mathop {\min }\limits_{{f_\theta }} \sum\limits_{i = 1}^N {hard({x_i},{y_i},{f_\theta }) \cdot \mathcal{L}({f_\theta }(\tilde x_i^K),{y_i})},
\end{equation}
where $hard({x_i},{y_i},{f_\theta })$ is the weight for each AE. 

Figure \ref{fig1} illustrates the pipeline of our proposed HAM framework. HAM judges an AE as easy or hard in terms of whether the AE crosses the decision boundary in $M$ attack iterations. Hard AEs are utilized in the subsequent training procedure, while easy AEs will be discarded. Different weights are allocated to each hard AE when calculating the loss function. To further save the computational cost of HAM, an early-dropping mechanism is applied to the easy AEs, which prevents them from the following training. The early-dropped easy AEs have zero weights in Equation \ref{Eq.3}.

\begin{figure}[!t]
\centering
\includegraphics[scale=0.47]{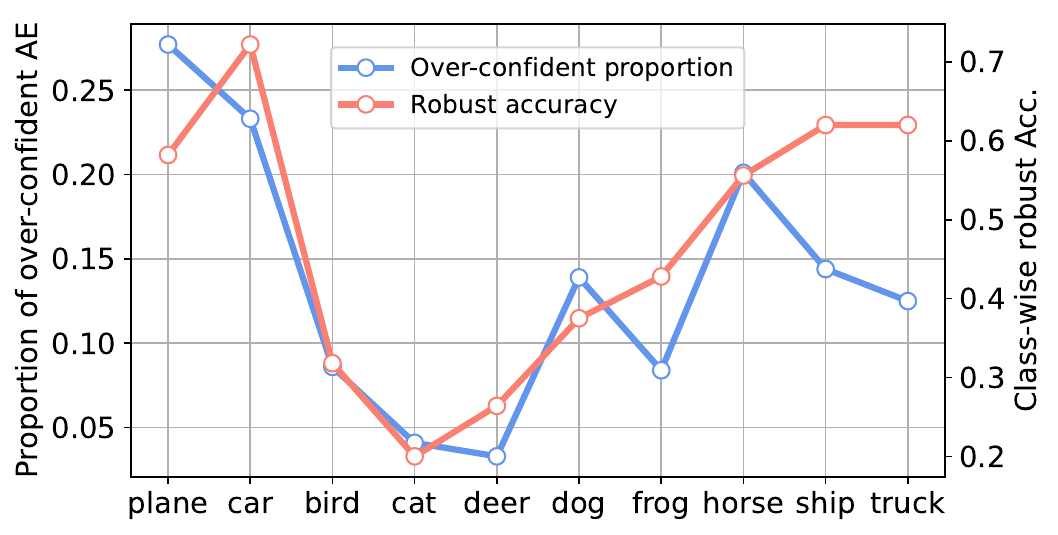} 
\caption{Class-wise over-confident AE proportion (blue line) and robustness (red line) of PGD-AT on CIFAR-10 dataset. 
Classes with higher over-confidence AE proportion have higher robustness in most cases.
}
\label{fig10}
\end{figure}

\subsection{Adversarial confidence overfitting}\label{Sec.3.3}
HAM applies sample mining techniques in AT to mitigate the adversarial confidence overfitting issue. Before moving on, we identify the existence of the adversarial confidence overfitting problem in AT by conducting a preliminary experiment as illustrated in Figure \ref{fig2}.
The sample points above the diagonal line represent the AEs with higher confidence than their corresponding clean examples (i.e., over-confident AEs). These results indicate that a large proportion of over-confident AEs will be produced during AT, which may lead to the unfairness problem in traditional AT methods. The dominance of over-confident AEs indicates that traditional AT methods pay more attention to the already-robust AEs (or easy AEs defined in this paper) while ignoring the more challenging AEs (hard AEs) that deserve more attention.  

The adversarial confidence overfitting issue also relates to the class-wise unfairness problem of AT. We can see in Figure \ref{fig10} that the proportion of over-confident AEs varies significantly across different classes. For instance, the proportions of over-confident AE of the class cat ($4.1\%$) and deer ($3.3\%$) are much lower compared to the class airplane ($27.7\%$) and automobile ($23.3\%$). This class-wise AE confidence disparity will make AT pay more attention to the classes with higher over-confidence AE ratios and less attention to other classes, resulting in the class-wise robust unfairness problem. \textcolor{black}{We additionally plot the class-wise robustness results in Figure \ref{fig10} to support this claim. It can be observed that the classes with higher over-confident AE proportions also have relatively higher robust accuracy.} This paper proposes HAM to mitigate the adversarial confidence overfitting issue, which reweights hard AEs and prevents AT from overfitting to easy AEs.

\textcolor{black}{We conduct experiments to further study the relationship between overconfident AEs used in the AT and the final fairness performance of the robust model.} We observe in Table \ref{Tab.1} that increasing the drop rate of over-confident AEs does not necessarily result in a significant fairness performance decline. These results indicate that not all AEs contribute to maintaining model fairness. Conversely, excluding over-confident AEs has the potential to improve model fairness. This insight motivates us to design HAM in a sample-mining way to mitigate the adversarial confidence overfitting issue, thereby improving robust fairness. 

\begin{table}[!t]
\centering
\begin{tabular}{c||ccccccc}
\toprule
Drop rate (\%)   & 0(AT)   & 10    & 20    & 30    \\
\midrule
Worst Std. (\%)  & 33.0 & 31.6 & 33.0 & 30.4  \\
Worst Rob. (\%) & 85.8 & 82.8 & 85.0 & 85.2 \\
\bottomrule
\end{tabular}
\caption{\textcolor{black}{The error rate of worst-class standard and robust results of training with confidence-wise AEs drop.}}
\label{Tab.1}
\end{table}

\subsection{Adaptive hard adversarial example mining}\label{Sec.3.4}
HAM mainly consists of an easy AE early-dropping stage and a hard AE reweighting stage. Easy AEs are identified in the first stage to prevent them from wasting the computational budget. Hard AEs are utilized and reweighted in the second training stage. 

\noindent \textbf{Easy AE early-dropping.}
In the early-dropping stage, given the total number of PGD iterations $K$, HAM first identifies easy AEs with $M$-step PGD ($M<K$). AEs that fail to cross the decision boundary within the $M$ PGD attack steps are identified as easy AEs, which will be dropped in this training epoch and will not participate in the following $(K-M)$-step AE generation process. The easy AE early-dropping mechanism saves the computational  budget and prevents AT from paying too much attention to the already-robust easy AEs.

\noindent \textbf{Hard AE reweighting.}
 HAM reweights the hard AEs in terms of the following metric:
\begin{equation}\label{Eq.4}
hard({x_i},{y_i},{f_\theta }) \!=\! \left\{  \begin{array}{l}
\!\omega (\mathop {\max }\limits_{1 \le j \le K} {\left\| {\Delta {f_\theta }(\tilde x_i^j)} \right\|_1}),{\rm{ }}{f_\theta }(\tilde x_i^M) \ne {y_i} \\\\

\!0,{\rm{ }}{f_\theta }(\tilde x_i^M) = {y_i}   \\

\end{array}  \right. \\
\end{equation}
where $\Delta {f_\theta }(\tilde x_i^j) = {f_\theta }(\tilde x_i^{j + 1}) - {f_\theta }(\tilde x_i^j)$ is the logits variant between two adjacent attack steps, $\omega( \cdot )$ is a monotonically increasing function $\omega (z) = sigmoid ( {z + \lambda } )$ with hyperparameter $\lambda$, ${\tilde x_i^j}$ represents the $M$-step AE used to distinguish the easy AE and hard AE. The AEs that fail to cross the decision boundary (${f_\theta }(\tilde x_i^M) = {y_i}$) will be identified as easy AEs and will be assigned zero weights.

\begin{figure}[!t]
\centering
\includegraphics[scale=0.35]{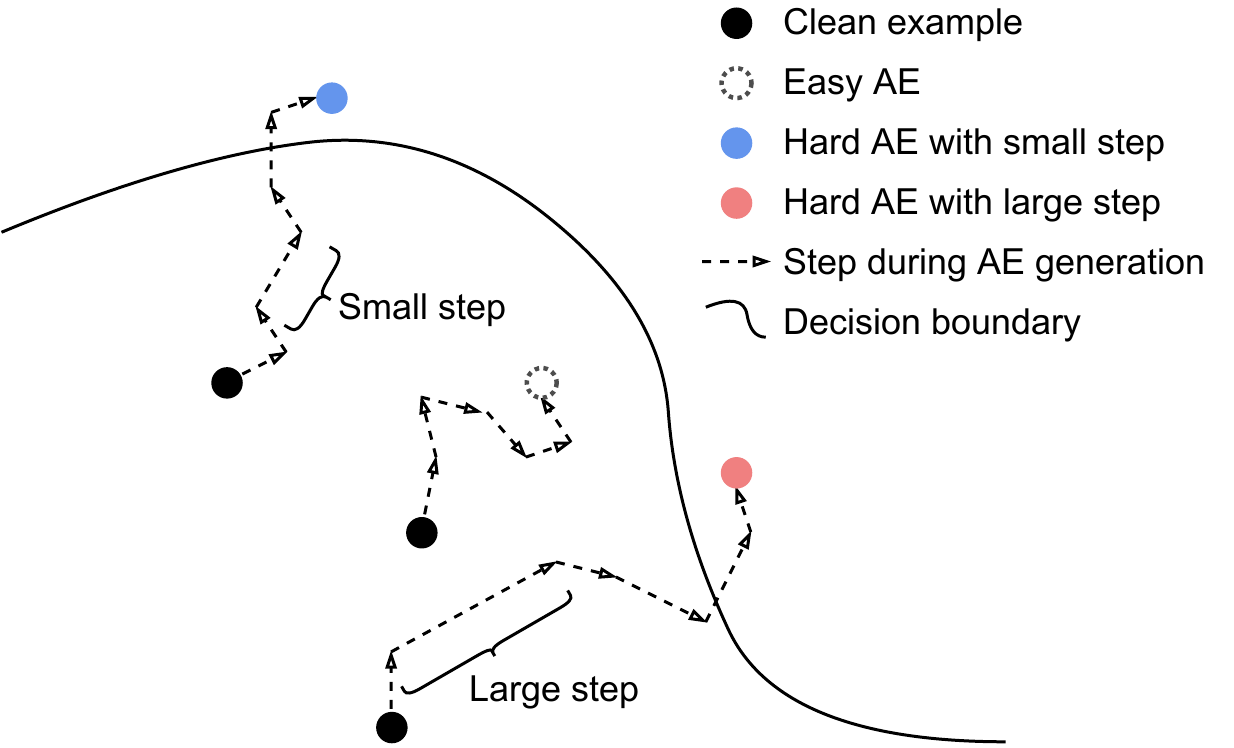} 
\caption{The intuition behind HAM.}
\label{fig4}
\end{figure}

The hard AEs with non-zero weights are utilized in the adversarial training. As illustrated in Figure \ref{fig4}, hard AEs with larger maximum adversarial step-sizes (adversarial step-sizes denotes the logits variants between two adjacent adversarial attack steps) will be assigned with larger weights, and vice versa. We summarize the proposed algorithm in Algorithm \ref{alg:algorithm}. In addition, \textcolor{black}{HAM is a general technique that can be integrated with other AT-based methods to mitigate the adversarial confidence overfitting issue and thus offers advantages in terms of both fairness and computational cost.}

\begin{algorithm}[!htb]
\caption{Pseudo-code of HAM}
\label{alg:algorithm}
\textbf{Input}: Network ${f_\theta }$, data $S = \{ ({x_i},{y_i})\} _{i = 1}^n$, batch size ${n_{bs}}$, number of batches ${n_b}$, learning rate $\eta $, training epochs $T$, whole attack step $K$, early-dropping step $M$\\
\textbf{Output}: Robust model ${f_\theta }$
\begin{algorithmic}[1] 
\FOR{epoch = $1, \cdots ,T$}
\STATE Sample a mini batch $\{ ({x_i},{y_i})\}_{i = 1}^{{n_{bs}}}$ from $S$
\FOR{mini-batch = $1, \cdots ,{n_b}$}
\STATE Generate $\{\tilde x_i^M\}_{i=1}^{n_{bs}}$ with $M$-step PGD
\STATE Construct hard AE set: $H_M = \{\tilde x_j^M: f_\theta (\tilde x_j^M) \neq {y_j}\}$
\STATE Finish hard AE generation with $(K-M)$ step PGD:   $H_K = \{\tilde x_j^K: \tilde x_j^M \in H_M\}$
\STATE Update $hard({x_i},{y_i},{f_\theta }), i=1,\cdots,n_{bs}$ by Equation \ref{Eq.4}
\STATE Update $\theta$ with SGD by Equation \ref{Eq.3}
\ENDFOR
\ENDFOR
\end{algorithmic}
\end{algorithm}
\noindent\textbf{Fairness benefits of HAM.} 
As has been discussed, the class-wise unfairness problem can be attributed to the phenomenon that AT pays more attention to the classes with higher ratios of over-confident AEs and less attention to other classes. As illustrated in Figure \ref{fig10}, the largely diverged proportion of over-confident easy AEs of each class and the robust overconfidence issue are the root causes of the robust unfairness problem. By mitigating the adversarial overconfidence issue with sampling mining techniques, HAM thus makes AT pay enough attention to the less-confident class and mitigates the robust unfairness problem. 

\begin{figure}[!t]
\centering
\includegraphics[scale=0.5]{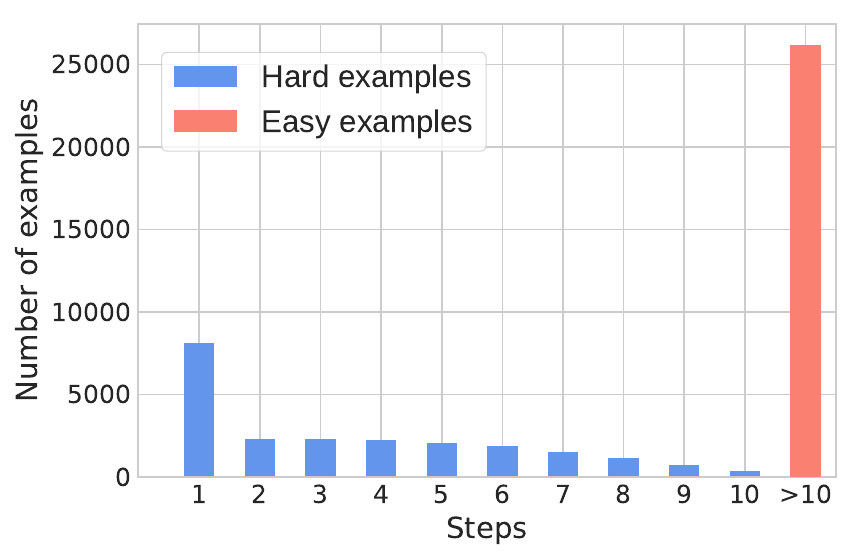} 
\caption{Number of examples with different minimal PGD steps for a successful attack.}
\label{fig3}
\end{figure}

\noindent\textbf{Efficiency benefits of HAM.} The early-dropping mechanism in HAM saves computational costs of AT while keeping robust fairness unchanged, which can be verified by the results in Figure \ref{fig3}.
\textcolor{black}{We can see that on the one hand, easy AE accounts for a large proportion ($53.6\%$) of the training dataset. Considering the computational cost of AT mainly concentrates on the AE generation stage, dropping these easy AEs thus significantly saves the computational budget. On the other hand, most of ($75\%$) the AEs that fail to attack the model before the first $M (=3)$ steps, cannot successfully attack the model in the end. This indicates that the early-dropping mechanism can detect easy AEs with a low false positive rate, which in turn does not compromise the final model's robustness. 
 As a result, early-dropping easy AEs will make a significant improvement in the efficiency of AT. More experimental evidence supporting the efficiency advantage of HAM is provided in Section \ref{Sec.4}.}

\section{Experiments}\label{Sec.4}
To evaluate the advantages of the proposed HAM in improving AT fairness and robustness, and saving training time, we perform several experiments on CIFAR-10, SVHN \cite{Netzer2011ReadingDI} and Imagenette, three widely-adopted image classification datasets for robust learning. Section \ref{Sec.4.2} evaluates the fairness improvement of HAM models.  Section \ref{Sec.4.3} presents the efficiency results of HAM. Parameter sensitivity experimental results are shown in Section \ref{Sec.4.4}.

\begin{table*}[!htb]
\centering
\resizebox{\linewidth}{!}{
    \begin{tabular}{l||cc|cc|cc}
    \toprule
    Method   & Avg. Std. ($\downarrow$) & Worst Std. ($\downarrow$) & Avg. Bndy. ($\downarrow$) & Worst Bndy. ($\downarrow$) & Avg. Rob. ($\downarrow$) & Worst Rob. ($\downarrow$) \\
    \midrule
    PGD-AT          & 15.38	&	33.00	&	41.24 	&	57.90 	&	56.62 	&	85.50            \\
    FRL(Reweight, 0.05)          & 16.45 	&	31.30 	&	39.78 	&	54.60 	&	56.23 	&	81.60            \\
    FRL(Reweight, 0.07)          & 16.59  	&	30.90  	&	39.78  	&	53.80  	&	56.37  	&	80.10             \\
    FRL(Reweight+Remargin, 0.05)          & 19.49 	&	32.50 	&	33.96 	&	42.10 	&	53.45 	&	72.80            \\
    FRL(Reweight+Remargin, 0.07)          & 18.56  	&	32.90  	&	35.37  	&	46.10  	&	53.93  	&	75.50             \\
    BAT          & \textbf{13.21}  	&	\textbf{25.80}  	&	36.63  	&	53.20  	&	49.84  	&	75.90             \\
    FAT          & 15.26  	&	31.50  	&	41.11  	&	58.40  	&	56.37  	&	83.90             \\
    \textbf{HAM(Ours)}        & 17.06  	&	28.90  	&	\textbf{30.99}	&	\textbf{35.30}	&	\textbf{48.05}	&	\textbf{64.20}           \\
    \bottomrule
    \end{tabular}
}
\caption{Error (\%) of average \& worst-class standard, boundary and robust for our HAM and other methods on CIFAR-10. The best results are in \textbf{bold}.}
\label{Tab.7}
\end{table*}

\begin{table*}[!htb]
\centering
%
\resizebox{\linewidth}{!}{
\begin{tabular}{l||cc|cc|cc}
\toprule
Method   & Avg. Std. ($\downarrow$) & Worst Std. ($\downarrow$) & Avg. Bndy. ($\downarrow$) & Worst Bndy. ($\downarrow$) & Avg. Rob. ($\downarrow$) & Worst Rob. ($\downarrow$) \\
\midrule
PGD-AT          & 7.82	&	12.17	&	41.31 	&	55.06 	&	49.13 	&	65.90            \\
FRL(Reweight, 0.05)          & 7.60 	&	12.63 	&	41.63 	&	56.86 	&	49.23 	&	65.90             \\
FRL(Reweight, 0.07)          & 8.22  	&	16.23  	&	40.40  	&	54.09  	&	48.62  	&	63.73             \\
FRL(Reweight+Remargin, 0.05)          & 7.53 	&	10.92 	&	41.06 	&	53.37 	&	48.59 	&	61.98            \\
FRL(Reweight+Remargin, 0.07)          & 9.30  	&	14.51  	&	39.37  	&	52.16  	&	48.67  	&	63.31             \\
BAT          & \textbf{5.03}  	&	\textbf{7.42}  	&	\textbf{29.25}  	&	53.43  	&	\textbf{34.28}  	&	57.40             \\
FAT          & 10.08  	&	16.26  	&	48.21  	&	59.33  	&	58.30  	&	75.60             \\
\textbf{HAM(Ours)}        & 8.26  	&	11.62  	&	30.06	&	\textbf{35.66}	&	38.33	&	\textbf{47.04}           \\
\bottomrule
\end{tabular}
}
\caption{Error (\%) of average \& worst-class standard, boundary and robust for our HAM and other methods on SVHN.}
\label{Tab.8}
\end{table*}

\begin{table*}[!htb]
\centering
%
\resizebox{\linewidth}{!}{
\begin{tabular}{l||cc|cc|cc}
\toprule
Method   & Avg. Std. ($\downarrow$) & Worst Std. ($\downarrow$) & Avg. Bndy. ($\downarrow$) & Worst Bndy. ($\downarrow$) & Avg. Rob. ($\downarrow$) & Worst Rob. ($\downarrow$) \\
\midrule
PGD-AT & 29.17 & 40.93 & 35.31 & 43.14 & 64.48 & 79.23 \\
FRL(Reweight, 0.05) & 29.94  & 41.28  & 35.38  & 45.17  & 65.32  & 78.75  \\
FRL(Reweight, 0.07) & 31.40  & 38.90  & 34.99 & 45.43 & 66.40  & 78.75 \\
FRL(Reweight+Remargin, 0.05) & 32.07 & \textbf{38.04} & 33.59 & 45.93 & 65.66 & 76.61 \\
FRL(Reweight+Remargin, 0.07) & 30.98 & 39.37 & 34.05 & 43.14 & 65.03 & 76.42 \\
BAT & 32.96 & 61.39 & \textbf{30.89} & 44.67 & 63.85 & 88.08 \\
FAT & \textbf{27.97} & 39.37 & 35.46 & 45.17 & 63.44 & 80.05 \\
\textbf{HAM(Ours)} & 29.40  & 41.19 & 31.59 & \textbf{39.79} & \textbf{60.99} & \textbf{73.03} \\
\bottomrule
\end{tabular}
}
\caption{Error (\%) of average \& worst-class standard, boundary and robust for our HAM and other methods on Imagenette.}
\label{Tab.9}
\end{table*}

\subsection{Experimental setup}\label{Sec.4.1}
\textcolor{black}{All the following experiments are performed on an Ubuntu 20.04 Operating System with Intel Xeon Gold 6226R CPUs and RTX 3090 GPUs.  When testing the efficiency of different AT methods, we use a single GPU for a fair comparison. The deep learning framework we use is PyTorch 1.9. We list the hyper-parameter settings on each dataset as follows.}

\textbf{CIFAR-10 \& Imagenette}.
The experiments on CIFAR-10 and Imagenette share the same experimental settings. For all the AT methods compared in this section, we trained PreActResNet-18 \cite{he2016identity} for 120 epochs. The batch size is set to 128. The optimizer is SGD with a momentum factor of 0.9 and a weight decay factor of $2 \times {10^{ - 4}}$. The initial learning rate is 0.1. At the 60$th$, 90$th$, and 110$th$ epochs, the learning rate is decayed to 10\%, 1\%, and 0.5\% of the initial learning rate, respectively. We adopt the commonly used image pre-processing methods and augmentations. Images on CIFAR-10 are normalized to $[0,1]$ and augmented with random crop and random flip. We use the $l_{\infty}$-norm PGD attack with the perturbation budget of $8/255$ to generate adversarial examples. 10-step PGD is used in the training stage and 20-step PGD is used in the test stage, which is consistent with previous works \cite{Pang2021BagOT,Rice2020OverfittingIA,Pang2020BoostingAT,Bai2021RecentAI}. The step size of the PGD attack is set to $2/255$. Our HAM method is started only after the 50th epoch, and the early-dropping hyper-parameter is set to $M = 3$. 

\textbf{SVHN}.
We only list the settings used on SVHN, which are different from those used on CIFAR-10. The initial learning rate is 0.01. The early-dropping hyper-parameter of HAM is set to $M = 5$. All the other settings remain the same as on CIFAR-10.

\textbf{Baseline method.}
We calculate the fairness and efficiency and compare our HAM with traditional PGD-AT (denoted as AT in the following) as well as several state-of-the-art methods designed for fairness improvement, including FRL \cite{Xu2021ToBR}, BAT \cite{sun2022improving}, and FAT \cite{ma2022tradeoff}. All the baseline methods follow the experimental settings in their original paper for a fair comparison.

\textcolor{black}{
\textbf{Evaluation metric.} 
We use the same fairness metrics as the previous studies \cite{Xu2021ToBR, sun2022improving}.
Standard error (Avg.Std.) and boundary error (Avg.Bndy.) correspond to the average error rates on misclassified clean examples and adversarial examples, respectively, and robust error is the sum of the Avg.Std. and the Avg.Bndy. The worst error (Worst Std., Worst Bndy. and Worst Rob.) corresponds to the largest error rate per class and the larger worst error indicates worse fairness. We report the worst error on both clean test examples (Worst Std.) and adversarial examples (Worst Bndy.) to comprehensively evaluate the fairness of the AT model. 
}

\subsection{Fairness of HAM}\label{Sec.4.2}
This section reports the experimental results that evaluate the fairness of HAM. We report the same fairness metric as the previous fair AT works \cite{Xu2021ToBR, sun2022improving} for a fair comparison. Two variants of FRL, reweight and remargin are adopted in this experiment. Besides, we further report the standard variance of class-wise robustness of each AT method as a supplement.
The results on CIFAR-10, SVHN, and Imagenette are reported in Table \ref{Tab.7}, Table \ref{Tab.8} and Table \ref{Tab.9}, respectively. We can see that the class-wise robust fairness performance of HAM significantly outperforms previous SOTA methods on all three datasets, achieving the lowest Worst Rob. We analyze the experimental results on each dataset as follows.

\begin{figure}[!t]
\centering
\includegraphics[scale=0.33]{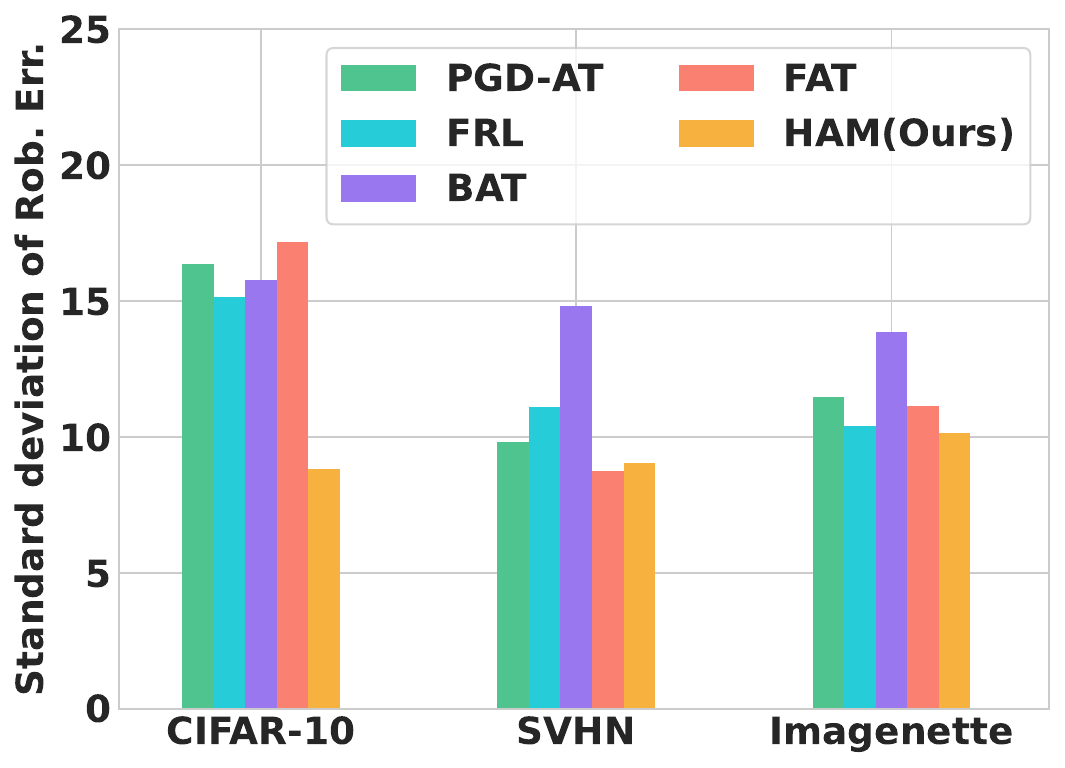} 
\caption{\textcolor{black}{The standard deviation of the robust accuracy of each AT model on the three datasets. A lower standard deviation means better robust fairness. Our HAM outperforms every compared method on almost every dataset.} 
}
\label{fig11}
\end{figure}

\textbf{CIFAR-10.} \textcolor{black}{
As shown in Table \ref{Tab.7}, HAM significantly reduces the worst boundary error, and robust error of PGD-AT by 22.6\%, and 21.3\%. Compared to SOTA FRL and BAT (best Worst Bndy. and Rob. results) methods, the improvements are 6.8\%, 8.6\%, and 17.9\%, 11.7\%, respectively. In addition, HAM performs better than PGD-AT, FRL, BAT, and FAT on average boundary error and robust error.}

\textbf{SVHN.} 
Similar improvements can be observed on SVHN dataset, as shown in Table \ref{Tab.8}. Compared to PGD-AT, HAM achieves 19.4\% and 18.8\% reduction to the worst boundary and worst robust error. 
HAM also significantly improves BAT by 17.7\% and 10.3\%. And HAM performs better than FRL (best one) regarding the worst boundary error and robust error with 17.7\% and 14.9\%. 
In terms of average boundary error and robust error, HAM improves PGD-AT, FRL, and FAT methods by a large margin and achieves comparable performance to BAT. These results highlight that HAM improves the class-wise robust fairness while not sacrificing the average robust performance.

\textcolor{black}{
\textbf{Imagenette.} 
As with the previous two datasets, HAM performs best on the worst boundary error and the worst robustness error of all the methods in our evaluation, as shown in Table \ref{Tab.9}. Compared to the standard PGD-AT, HAM reduces the Worst Bndy. error by 3.3\% and the Worst Rob. error by 6.2\%. Besides, our HAM does not severely degrade the average robustness performance, the average robustness performance drops of HAM compared to FRL, BAT, and FAT are only 3.1\% on the Worst Std. error, 0.7\% on the Avg. Bndy. error, and 1.4\% on the Avg. Std. error, which are trivial compared to our fairness improvement. 
}

\textcolor{black}{
Current evaluation metrics assess the fairness of robust models from the perspective of the average and worst category. However, they do not account for differences and variances in robustness across categories, which is crucial for a comprehensive fairness evaluation. Therefore, we propose to count the standard deviation of the model robustness across categories as an additional metric. As can be seen in Figure \ref{fig11}, the standard deviation of robustness between categories of HAM is significantly lower than that of other methods, further demonstrating the effectiveness of HAM.
}

\begin{figure*}[!t]
\centering
\subfloat[CIFAR-10]{\includegraphics[scale=0.295]{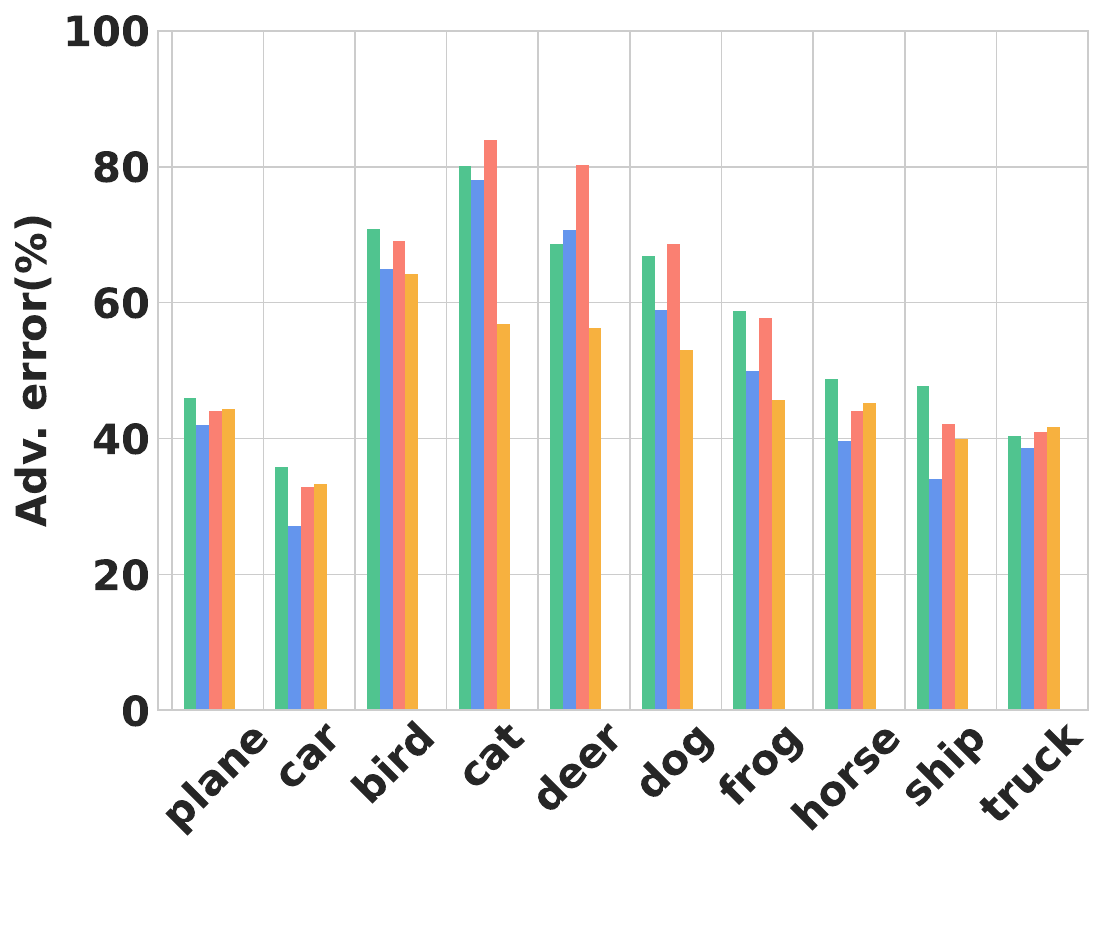}\label{Fig.9.1}}~
\subfloat[SVHN]{\includegraphics[scale=0.295]{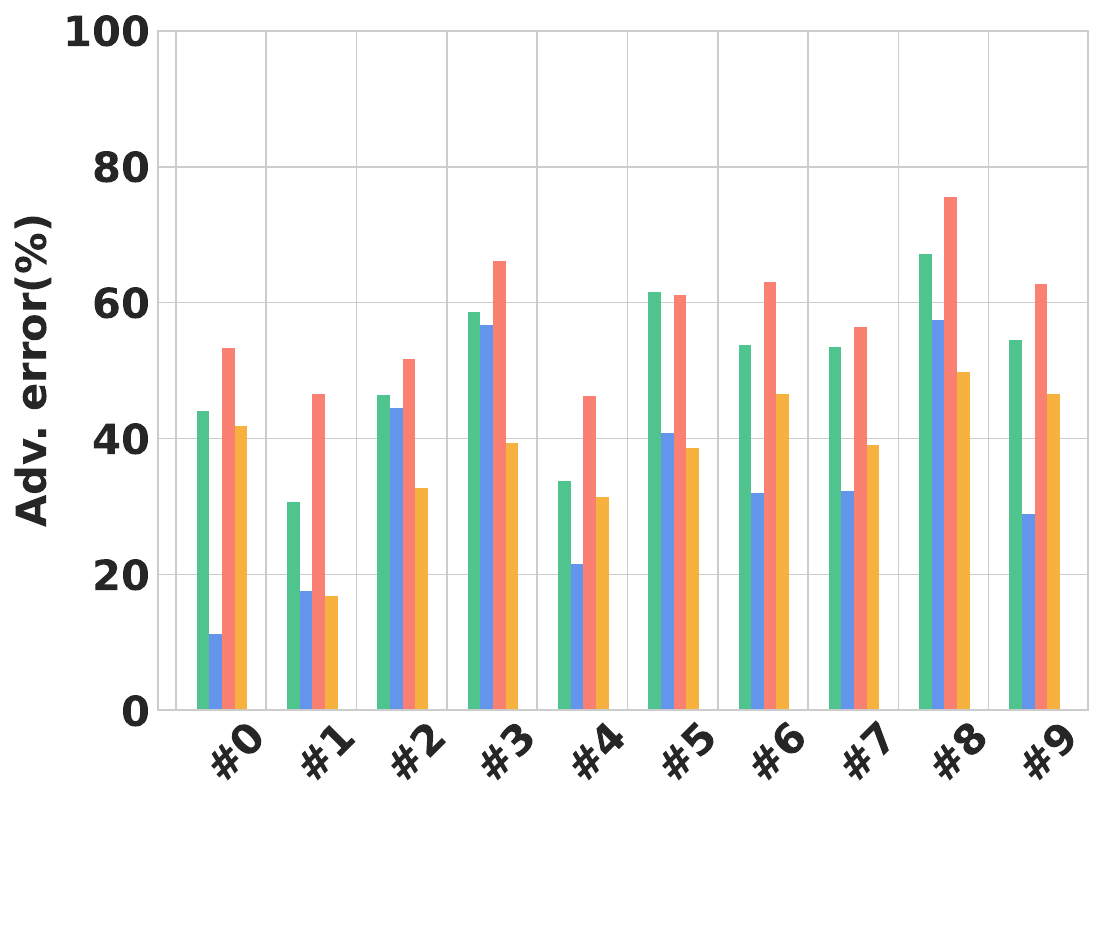}\label{Fig.9.2}}~
\subfloat[Imagenette]{\includegraphics[scale=0.295]{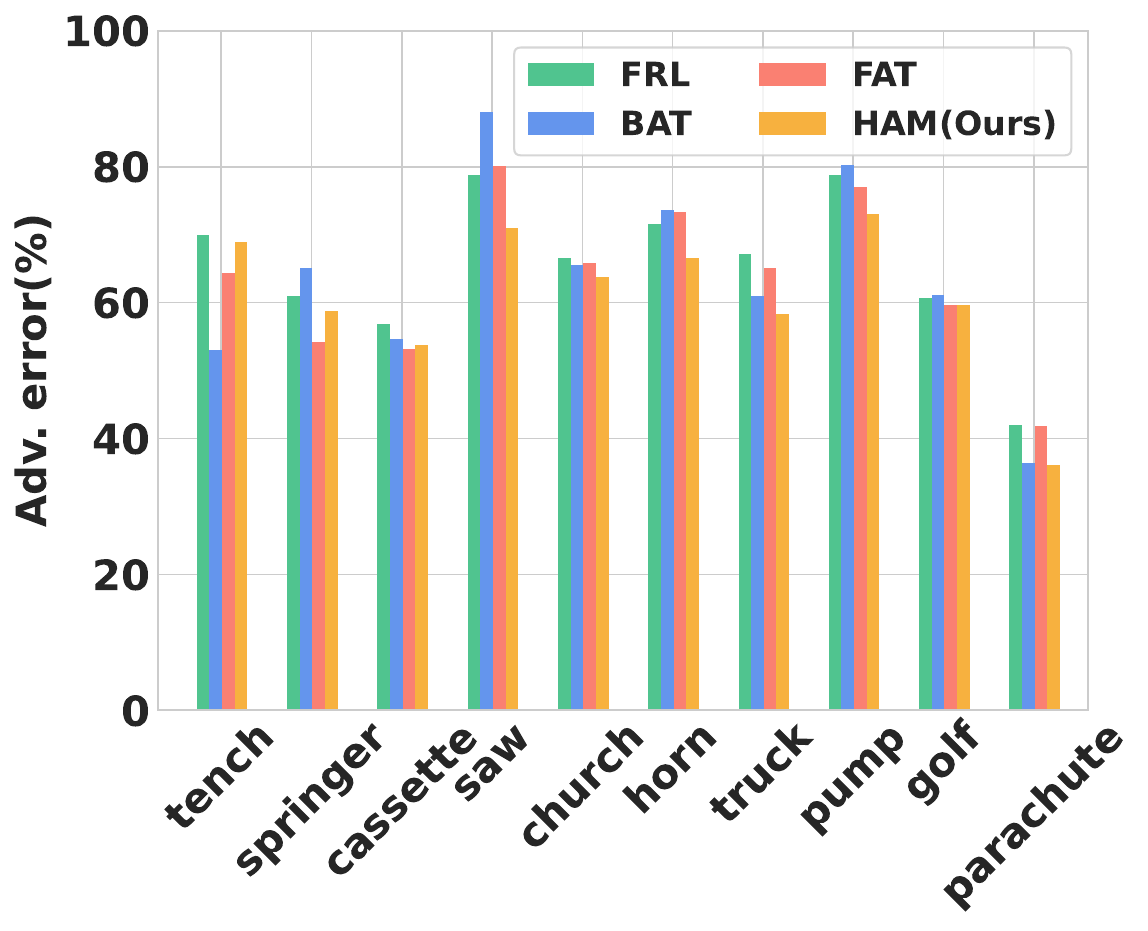}\label{Fig.9.3}}
\caption{Fairness: Adversarial error under PGD-20 attack in each class of three datasets.}
\label{Fig.9}
\end{figure*}

In addition, we report the robust performance of AT methods on each class in Figure \ref{Fig.9}, we can see that HAM outperforms other methods on several classes, especially those that get inferior in fairness evaluation when trained with baseline AT methods, e.g. \textit{cat, deer, dog, and frog} (CIFAR-10) and \textit{\#1, \#2, \#3, \#7, and \#8} (SVHN).


\begin{table}[!t]
\centering
\resizebox{\columnwidth}{!}{
\begin{tabular}{l||lll}
\toprule
Algorithm & Worst Bndy. ($\downarrow$)       & Worst Rob. ($\downarrow$)     & Training time ($\downarrow$) \\
\midrule
AT	&	57.90 	&	85.50 	&	142     \\
BAT	&	53.20 (-4.7) 	&	75.90 (-9.6) 	&	498 (+250\%) 	    \\
Fast AT	&	48.70 (-9.2) 	&	86.30 (+0.8) 	&	\textbf{16 (-88\%)} 	    \\
HAM (Ours)	&	\textbf{35.30 (-22.6)} 	&	\textbf{64.20 (-21.3)} 	&	78 (-45\%) 	   \\
\bottomrule
\end{tabular}}
\caption{\textcolor{black}{Comparing the fairness and efficiency of HAM with baseline AT methods. The best results are in \textbf{bold}. We report the improvement magnitude of each method compared with PGD-AT (denoted as AT) in the brackets.}}
\label{Tab.5}
\end{table}

\begin{table}[!t]
\centering
\resizebox{\columnwidth}{!}{
\begin{tabular}{l||lll}
\toprule
Algorithm & Worst Bndy. ($\downarrow$)       & Worst Rob. ($\downarrow$)     & Training time ($\downarrow$) \\
\midrule
AT	&	57.9 	&	85.5 	&	142     \\
HAM (Ours)	&	35.3 (-22.6) 	&	64.2 (-21.3) 	&	\textbf{78 (-45\%)} 	   \\
MART	&	49.5 (-8.4) 	&	75.5 (-10.0) 	&	131 (-8\%) 	    \\
MART-HAM	&	\textbf{21.9 (-36.0)} 	&	\textbf{51.0 (-34.5)} 	&	99 (-30\%) 	    \\
\bottomrule
\end{tabular}}
\caption{The robust fairness and training efficiency of HAM when being combined with one of the SOTA AT methods, MART. The best results are in \textbf{bold}.} 
\label{Tab.11}
\vspace{-0.2cm}
\end{table}

\subsection{Efficiency and extendibility of HAM}\label{Sec.4.3}
The training time of HAM and other AT methods on CIFAR-10 is reported to verify the efficiency advantage of HAM. We also verify the extendibility of HAM by combining it with one of the SOTA AT methods, MART \cite{Wang2020ImprovingAR}. 

The training time in Table \ref{Tab.5} represents the time (seconds) it takes PreActResNet-18 to train an epoch on CIFAR-10. We can see that accelerated AT (Fast AT) greatly saves training time while worsening the AT fairness. On the contrary, fair AT (BAT) improves fairness while sacrificing training efficiency, making the already inefficient AT slower. HAM improves both results simultaneously.

In addition, HAM is a generic method, which means it can be easily combined with other AT methods to improve fairness and efficiency. Table \ref{Tab.11} shows the results of HAM being extended to MART. HAM reduces the Worst Bndy. error and Worst Rob. error of MART by 36.0\% and 34.5\%, respectively, while also reducing training time by 30\%.

\subsection{Ablation study}\label{Sec.4.4}
\begin{figure}[!t]
\centering
\subfloat[CIFAR-10]{\includegraphics[scale=0.285]{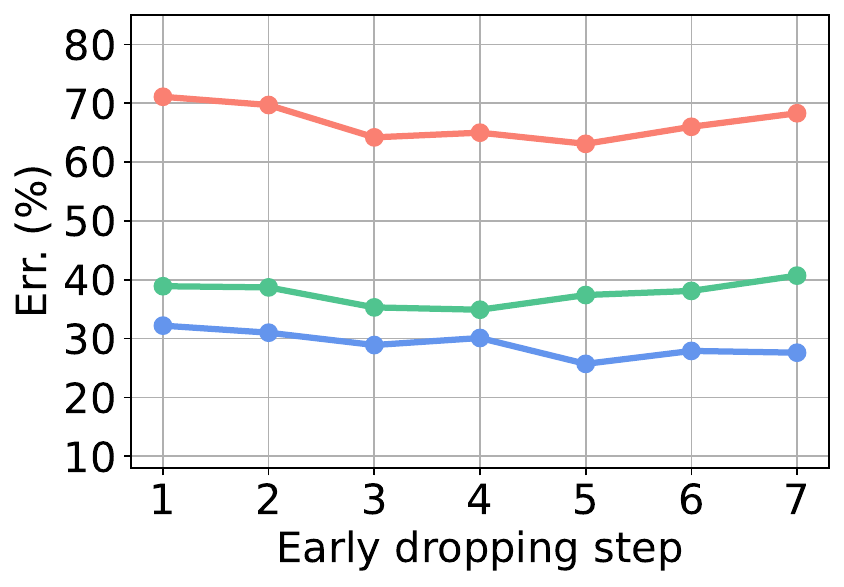}\label{Fig.5.1}}~
\subfloat[SVHN]{\includegraphics[scale=0.285]{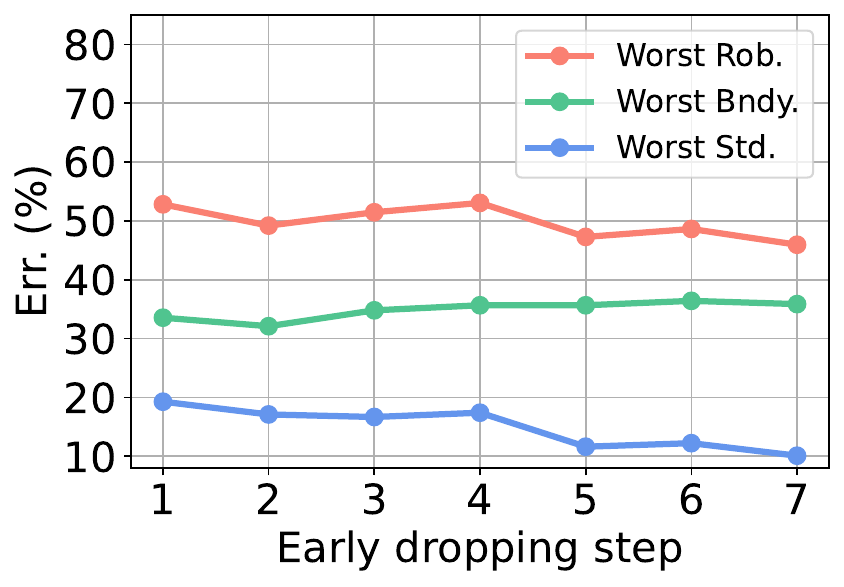}\label{Fig.5.2}}
\caption{Influence of early dropping step on fairness.}
\label{Fig.5}
\vspace{-0.4cm}
\end{figure}

\noindent \textbf{Hyper-parameter sensitivity.} The early-dropping strategy benefits the time saving, while the attack step $M$ of early-dropping directly affects the model performance.
Here, we explore the relationship between different steps and model fairness.
It can be seen in Figure \ref{Fig.5} that on both two datasets, too few steps can negatively affect the final fairness of the model.
As the step increases above 3, the negative effects become small enough.
Based on this result, selecting step from 3 to 5 can not only satisfy the model fairness but also significantly save training time.

\begin{figure}[!t]
\centering
\subfloat[CIFAR-10]{\includegraphics[scale=0.285]{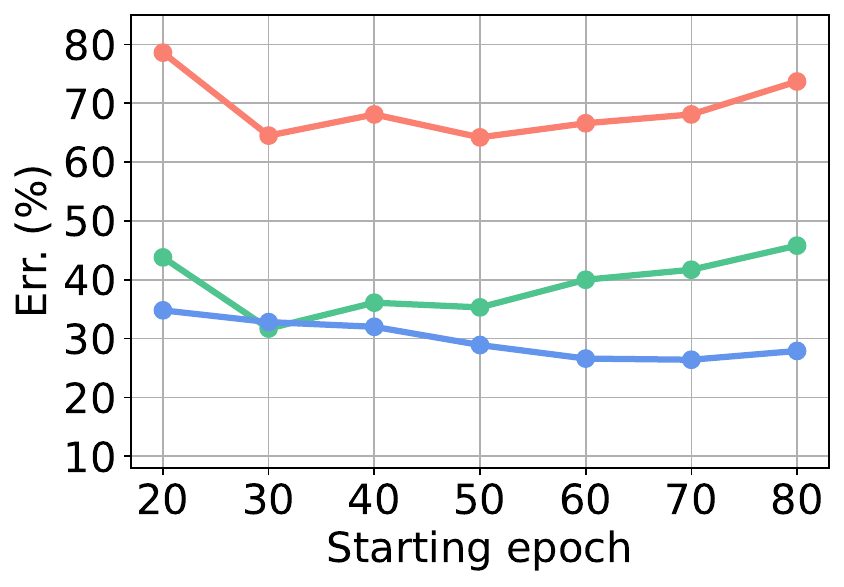}\label{Fig.6.1}}~
\subfloat[SVHN]{\includegraphics[scale=0.285]{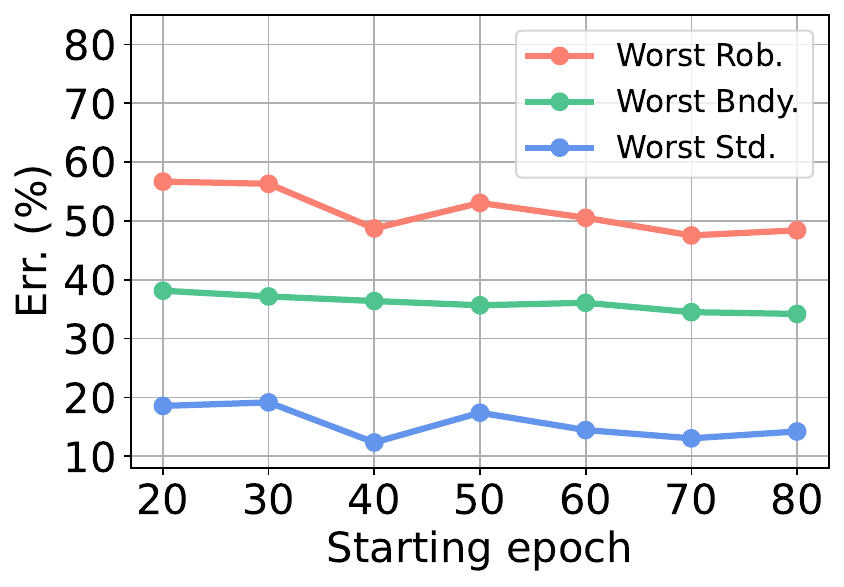}\label{Fig.6.2}}
\caption{Influence of begin epoch on fairness.}
\label{Fig.6}
\vspace{-0.4cm}
\end{figure}

HAM is not carried out from the very beginning, otherwise, it will lead to the fluctuation of the model training and the problem of difficult convergence.
In this part, we fix $M$ and analyze the selection of starting epoch.
Figure \ref{Fig.6} shows that starting epoch is appropriate within a range.
In a total of 120 epochs of training, when the learning rate is reduced to 10\% of the maximum learning rate in the 60$th$ epoch, the starting time between 40 and 60 will make a good effect.
Therefore, we conclude that a starting epoch slightly earlier than the first drop in the learning rate performs well.

\begin{table}[!t]
\centering
\resizebox{\columnwidth}{!}{
\begin{tabular}{c||ccccc|c}
\toprule
\multirow{2}{*}{Method} & \multicolumn{5}{c|}{Random drop rate (\%)}                         & \multirow{2}{*}{HAM} \\
                        & 0(AT)    & 10    & 20    & 30    & 40        &                       \\
\midrule
Worst Std.(\%)              & 33.00  & 34.20 & 32.20 & 33.50 & 33.40  & \textbf{28.90}                \\
Worst Bndy.(\%)             & 57.90 & 56.38 & 57.30 & 55.60 & 55.10   & \textbf{35.30}      \\
Worst Rob.(\%)             & 85.50 & 84.80 & 83.70 & 84.30 & 83.50   & \textbf{64.20}      \\
\bottomrule
\end{tabular}
}
\caption{Fairness comparison of HAM and training with a random drop on CIFAR-10.}
\label{Tab.10}
\end{table}

\begin{table}[!t]
\centering
\resizebox{\columnwidth}{!}{
\begin{tabular}{c||ccccc|c}
\toprule
\multirow{2}{*}{Method} & \multicolumn{5}{c|}{Random drop rate (\%)}        & \multirow{2}{*}{HAM} \\
                        & 0(AT)   & 10    & 20    & 30    & 40     &                       \\
\midrule
Worst Std.(\%)  &	12.17  	&	12.00 	&	11.62 	&	11.65 	&	12.17 	 	&	\textbf{11.62}    \\
Worst Bndy.(\%) &	55.06  	&	48.37 	&	54.09 	&	53.97 	&	55.00 	 	&	\textbf{35.66} \\
Worst Rob.(\%) &	65.90  	&	65.96 	&	64.81 	&	64.81 	&	65.84 	 	&	\textbf{47.04} \\
\bottomrule
\end{tabular}
}
\caption{Fairness comparison of HAM and training with a random drop on SVHN.}
\label{Tab.6}
\vspace{-0.4cm}
\end{table}

\noindent \textbf{Comparison with random dropping.} To further confirm the contribution of our HAM method in improving class-wise robust fairness, we compare the HAM and the AT with random dropping on CIFAR-10 (Table \ref{Tab.10}) and SVHN (Table \ref{Tab.6}). Random dropping means that we randomly drop AEs in the training procedure, rather than judging by $M$-step PGD attack as in HAM. We can see that the random dropping groups do not improve the fairness performance compared to the naive PGD-AT method, which highlights the contribution of our HAM method.

\section{Conclusion}\label{Sec.5}
This paper focuses on improving the robust fairness and efficiency of AT while still keeping the satisfying average robustness performance. We first reveal the adversarial confidence overfitting phenomenon that hinders the promotion of fairness. The empirical analysis motivates us to propose HAM, which mitigates the above issues by paying more attention to the hard AEs. The proposed HAM framework makes more use of hard AEs and drops the easy AEs, remarkably improving fairness and efficiency compared to several SOTA approaches on three datasets. 

{\small
\bibliographystyle{ieee_fullname}
\bibliography{paper}

\begin{thebibliography}{10}\itemsep=-1pt

\bibitem{Bai2021RecentAI}
Tao Bai, Jinqi Luo, Jun Zhao, Bihan Wen, and Qian Wang.
\newblock Recent advances in adversarial training for adversarial robustness.
\newblock In {\em IJCAI}, 2021.

\bibitem{Cui2019ClassBalancedLB}
Yin Cui, Menglin Jia, Tsung-Yi Lin, Yang Song, and Serge~J. Belongie.
\newblock Class-balanced loss based on effective number of samples.
\newblock {\em 2019 IEEE/CVF Conference on Computer Vision and Pattern
  Recognition (CVPR)}, pages 9260--9269, 2019.

\bibitem{girshick2014rich}
Ross Girshick, Jeff Donahue, Trevor Darrell, and Jitendra Malik.
\newblock Rich feature hierarchies for accurate object detection and semantic
  segmentation.
\newblock In {\em Proceedings of the IEEE conference on computer vision and
  pattern recognition}, pages 580--587, 2014.

\bibitem{Goodfellow2015ExplainingAH}
Ian~J. Goodfellow, Jonathon Shlens, and Christian Szegedy.
\newblock Explaining and harnessing adversarial examples.
\newblock {\em CoRR}, abs/1412.6572, 2015.

\bibitem{he2015spatial}
Kaiming He, Xiangyu Zhang, Shaoqing Ren, and Jian Sun.
\newblock Spatial pyramid pooling in deep convolutional networks for visual
  recognition.
\newblock {\em IEEE transactions on pattern analysis and machine intelligence},
  37(9):1904--1916, 2015.

\bibitem{he2016identity}
Kaiming He, Xiangyu Zhang, Shaoqing Ren, and Jian Sun.
\newblock Identity mappings in deep residual networks.
\newblock In {\em European conference on computer vision}, pages 630--645.
  Springer, 2016.

\bibitem{kim2021entropy}
Minseon Kim, Jihoon Tack, Jinwoo Shin, and Sung~Ju Hwang.
\newblock Entropy weighted adversarial training.
\newblock In {\em ICML 2021 Workshop on Adversarial Machine Learning}, 2021.

\bibitem{Krizhevsky2009LearningML}
Alex Krizhevsky.
\newblock Learning multiple layers of features from tiny images.
\newblock 2009.

\bibitem{Lin2020FocalLF}
Tsung-Yi Lin, Priya Goyal, Ross~B. Girshick, Kaiming He, and Piotr Doll{\'a}r.
\newblock Focal loss for dense object detection.
\newblock {\em IEEE Transactions on Pattern Analysis and Machine Intelligence},
  42:318--327, 2020.

\bibitem{ma2022tradeoff}
Xinsong Ma, Zekai Wang, and Weiwei Liu.
\newblock On the tradeoff between robustness and fairness.
\newblock In {\em Advances in Neural Information Processing Systems}, 2022.

\bibitem{Madry2018TowardsDL}
Aleksander Madry, Aleksandar Makelov, Ludwig Schmidt, Dimitris Tsipras, and
  Adrian Vladu.
\newblock Towards deep learning models resistant to adversarial attacks.
\newblock {\em ArXiv}, abs/1706.06083, 2018.

\bibitem{Netzer2011ReadingDI}
Yuval Netzer, Tao Wang, Adam Coates, A. Bissacco, Bo Wu, and A. Ng.
\newblock Reading digits in natural images with unsupervised feature learning.
\newblock 2011.

\bibitem{Pang2021BagOT}
Tianyu Pang, Xiao Yang, Yinpeng Dong, Hang Su, and Jun Zhu.
\newblock Bag of tricks for adversarial training.
\newblock {\em ArXiv}, abs/2010.00467, 2021.

\bibitem{Pang2020BoostingAT}
Tianyu Pang, Xiao Yang, Yinpeng Dong, Kun Xu, Hang Su, and Jun Zhu.
\newblock Boosting adversarial training with hypersphere embedding.
\newblock {\em ArXiv}, abs/2002.08619, 2020.

\bibitem{Rade2022ReducingEM}
Rahul Rade and Seyed-Mohsen Moosavi-Dezfooli.
\newblock Reducing excessive margin to achieve a better accuracy vs. robustness
  trade-off.
\newblock In {\em International Conference on Learning Representations}, 2022.

\bibitem{Rice2020OverfittingIA}
Leslie Rice, Eric Wong, and J.~Zico Kolter.
\newblock Overfitting in adversarially robust deep learning.
\newblock In {\em ICML}, 2020.

\bibitem{Shafahi2019AdversarialTF}
Ali Shafahi, Mahyar Najibi, Amin Ghiasi, Zheng Xu, John~P. Dickerson, Christoph
  Studer, Larry~S. Davis, Gavin Taylor, and Tom Goldstein.
\newblock Adversarial training for free!
\newblock In {\em NeurIPS}, 2019.

\bibitem{Shrivastava2016TrainingRO}
Abhinav Shrivastava, Abhinav~Kumar Gupta, and Ross~B. Girshick.
\newblock Training region-based object detectors with online hard example
  mining.
\newblock {\em 2016 IEEE Conference on Computer Vision and Pattern Recognition
  (CVPR)}, pages 761--769, 2016.

\bibitem{sun2022improving}
Chunyu Sun, Chenye Xu, Chengyuan Yao, Siyuan Liang, Yichao Wu, Ding Liang,
  XiangLong Liu, and Aishan Liu.
\newblock Improving robust fairness via balance adversarial training.
\newblock {\em arXiv preprint arXiv:2209.07534}, 2022.

\bibitem{sun2022towards}
Haipei Sun, Kun Wu, Ting Wang, and Wendy~Hui Wang.
\newblock Towards fair and robust classification.
\newblock In {\em 2022 IEEE 7th European Symposium on Security and Privacy
  (EuroS\&P)}, pages 356--376. IEEE, 2022.

\bibitem{Szegedy2014IntriguingPO}
Christian Szegedy, Wojciech Zaremba, Ilya Sutskever, Joan Bruna, D. Erhan,
  Ian~J. Goodfellow, and Rob Fergus.
\newblock Intriguing properties of neural networks.
\newblock {\em CoRR}, abs/1312.6199, 2014.

\bibitem{Toneva2019AnES}
Mariya Toneva, Alessandro Sordoni, R{\'e}mi~Tachet des Combes, Adam Trischler,
  Yoshua Bengio, and Geoffrey~J. Gordon.
\newblock An empirical study of example forgetting during deep neural network
  learning.
\newblock {\em ArXiv}, abs/1812.05159, 2019.

\bibitem{Wang2021ImbalancedAT}
Wentao Wang, Han Xu, Xiaorui Liu, Yaxin Li, Bhavani~M. Thuraisingham, and
  Jiliang Tang.
\newblock Imbalanced adversarial training with reweighting.
\newblock {\em ArXiv}, abs/2107.13639, 2021.

\bibitem{Wang2020ImprovingAR}
Yisen Wang, Difan Zou, Jinfeng Yi, James Bailey, Xingjun Ma, and Quanquan Gu.
\newblock Improving adversarial robustness requires revisiting misclassified
  examples.
\newblock In {\em ICLR}, 2020.

\bibitem{Wong2020FastIB}
Eric Wong, Leslie Rice, and J.~Zico Kolter.
\newblock Fast is better than free: Revisiting adversarial training.
\newblock {\em ArXiv}, abs/2001.03994, 2020.

\bibitem{Xu2021ToBR}
Han Xu, Xiaorui Liu, Yaxin Li, and Jiliang Tang.
\newblock To be robust or to be fair: Towards fairness in adversarial training.
\newblock In {\em ICML}, 2021.

\bibitem{ye2021amata}
Nanyang Ye, Qianxiao Li, Xiao-Yun Zhou, and Zhanxing Zhu.
\newblock Amata: An annealing mechanism for adversarial training acceleration.
\newblock In {\em Proceedings of the AAAI Conference on Artificial
  Intelligence}, volume~35, pages 10691--10699, 2021.

\bibitem{Zagoruyko2016WideRN}
Sergey Zagoruyko and Nikos Komodakis.
\newblock Wide residual networks.
\newblock {\em ArXiv}, abs/1605.07146, 2016.

\bibitem{Zhang2019YouOP}
Dinghuai Zhang, Tianyuan Zhang, Yiping Lu, Zhanxing Zhu, and Bin Dong.
\newblock You only propagate once: Accelerating adversarial training via
  maximal principle.
\newblock In {\em NeurIPS}, 2019.

\bibitem{Zhang2019TheoreticallyPT}
Hongyang~R. Zhang, Yaodong Yu, Jiantao Jiao, Eric~P. Xing, Laurent~El Ghaoui,
  and Michael~I. Jordan.
\newblock Theoretically principled trade-off between robustness and accuracy.
\newblock {\em ArXiv}, abs/1901.08573, 2019.

\bibitem{Zhang2021GeometryawareIA}
Jingfeng Zhang, Jianing Zhu, Gang Niu, Bo Han, Masashi Sugiyama, and M.
  Kankanhalli.
\newblock Geometry-aware instance-reweighted adversarial training.
\newblock {\em ArXiv}, abs/2010.01736, 2021.

\bibitem{Zheng2020EfficientAT}
Haizhong Zheng, Ziqi Zhang, Juncheng Gu, Honglak Lee, and Atul Prakash.
\newblock Efficient adversarial training with transferable adversarial
  examples.
\newblock {\em 2020 IEEE/CVF Conference on Computer Vision and Pattern
  Recognition (CVPR)}, pages 1178--1187, 2020.

\end{thebibliography}
}

\end{document}